\documentclass[11pt,letterpaper]{article}

\usepackage[T1]{fontenc}
\usepackage[utf8]{inputenc}
\usepackage{amsmath}
\usepackage{newtxtext,newtxmath}
\usepackage[
  letterpaper,
  top=0.88in,
  bottom=0.92in,
  left=0.95in,
  right=0.95in,
  headsep=0.24in
]{geometry}
\usepackage{booktabs}
\usepackage{tabularx}
\usepackage{array}
\usepackage{enumitem}
\usepackage{float}
\usepackage[section]{placeins}
\usepackage{etoolbox}
\usepackage{xcolor}
\usepackage{listings}
\usepackage{caption}
\usepackage{titlesec}
\usepackage{fancyhdr}
\usepackage{microtype}
\usepackage{xurl}
\usepackage{hyperref}

\definecolor{pulseink}{HTML}{111111}
\definecolor{pulserule}{HTML}{686868}
\definecolor{pulselink}{HTML}{343434}
\definecolor{pulsegray}{HTML}{F4F4F4}
\definecolor{pulsemuted}{HTML}{666666}

\hypersetup{
  colorlinks=true,
  linkcolor=pulseink,
  citecolor=pulseink,
  urlcolor=pulselink,
  pdftitle={PULSE: An Executable Contract Language for Spatiotemporal Knowledge Graph Engineering},
  pdfauthor={Dongxu Yang and Ziyi Liang},
  pdfsubject={Executable contracts for spatiotemporal knowledge graph engineering},
  pdfkeywords={knowledge graphs, executable contracts, spatiotemporal semantics, domain-specific language, GeoSPARQL}
}
\titleformat{\section}
  {\large\bfseries\color{pulseink}}{\thesection}{0.7em}{}
\titleformat{\subsection}
  {\normalsize\bfseries\color{pulseink}}{\thesubsection}{0.65em}{}
\titleformat{\subsubsection}
  {\normalsize\bfseries}{\thesubsubsection}{0.6em}{}
\titlespacing*{\section}{0pt}{1.8ex plus 0.5ex minus 0.2ex}{0.8ex}
\titlespacing*{\subsection}{0pt}{1.45ex plus 0.4ex minus 0.2ex}{0.55ex}
\titlespacing*{\subsubsection}{0pt}{1.2ex plus 0.3ex minus 0.2ex}{0.4ex}
\titlespacing*{\paragraph}{0pt}{1.0ex plus 0.3ex minus 0.1ex}{0.7em}

\fancypagestyle{plain}{%
  \fancyhf{}
  \fancyfoot[C]{\small\thepage}
  
}

\newcolumntype{Y}{>{\raggedright\arraybackslash}X}
\lstdefinelanguage{PULSE}{
  morekeywords={model,version,crs,region,entity,property,state,oneof,process,when,for,changes,constraint,must,inside,coveredBy,leaves,instance,observe,at,source,confidence,accuracy,scenario,assume,run,ask,while},
  sensitive=true,
  morestring=[b]",
  morecomment=[l]{//}
}
\setlist{nosep,leftmargin=1.5em}

\AtBeginEnvironment{thebibliography}{%
  \footnotesize
  \setlength{\itemsep}{0.1em}
  \setlength{\parskip}{0pt}
}
\begin{document}
\thispagestyle{plain}

\begin{center}
  {\LARGE\sffamily\bfseries\color{pulseink}
    PULSE: An Executable Contract Language for\\[0.15em]
    Spatiotemporal Knowledge Graph Engineering\par}
  \vspace{1.2em}
  {\large\bfseries Dongxu Yang\textsuperscript{*}\quad Ziyi Liang\par}
  \vspace{0.35em}
  % Deliberate arXiv deviation: use a text-only affiliation while the
  % DeepLethe institution icon is under review.
  {\normalsize\sffamily\bfseries
    \href{https://deeplethe.com/}{DeepLethe}\par}
  \vspace{0.2em}
  {\small
    \href{mailto:yangdongxu@deeplethe.com}{yangdongxu@deeplethe.com}\quad
    \href{mailto:liangziyi@deeplethe.com}{liangziyi@deeplethe.com}\par}
  \vspace{0.3em}
  {\small\textsuperscript{*}Corresponding author.\par}
  \vspace{0.35em}
  {\small\color{pulsemuted}Preprint, July 2026\par}
\end{center}

\vspace{0.35em}
\noindent\textcolor{pulserule}{\rule{\textwidth}{0.8pt}}
\vspace{-0.35em}

\begin{abstract}
Knowledge graph engineering often distributes accepted state, observations,
constraints, processes, and hypothetical scenarios across artifacts whose
combined execution contract remains external. We present PULSE, an
Object-Process-Methodology-inspired language that localizes four operational
roles and their write effects in one typed runtime. Here, modes denote
operational roles rather than modal or deontic logic. The implemented contract fixes
evidence non-overwrite, branch isolation, grounded multi-subject timers, guarded
state change, and declaration-ranked event ordering over time and space; an
external runner still decides whether evidence becomes an authoritative move.
GeoSPARQL, SOSA, and SHACL remain generated views. A core calculus gives an
effect-confinement lemma and six safety properties. Lean 4 checks kernel
analogues for positions, evidence, clocks, monitors, atomicity, and branch
source retention; 88 tests, 3,534 bounded checks, and 32 Lean/Python
runtime-kernel cases bound the implementation claim to the checked cases.
First-author implementations of a standards composition and a separate Sismic
statechart reproduce the tested cold-chain trace. Across
37,440 generated temporal traces, PULSE matches a separate workflow and
distinguishes ten single-field mutants. On the complete NOAA IBTrACS
\texttt{since1980} subset it agrees with GEOS and an event sweep on 1,476,290
transition-zone pairs, including 4,800 sampled and 12,831 duration-qualified
events. Project-specific GeoSPARQL probes measure interface coverage. Overall,
the results support contract localization, safety arguments, and trace parity
for the tested fragment; language superiority and usability remain outside the
evaluation.

\end{abstract}

\vspace{0.15em}
\noindent\textcolor{pulserule}{\rule{\textwidth}{0.35pt}}

\section{Introduction}

Knowledge graphs (KGs) describe entities and relations, but operational models
also record evidence, constrain states, react to time and location, and explore
alternatives. These concerns are commonly split across ontologies, observation
and shape graphs, process models, spatial stores, and simulators. Their
semantics then depend on artifact boundaries: observations are not accepted
updates, policies state obligations, scenarios must not mutate asserted state,
and duration-qualified departures are neither OWL axioms nor static SHACL
violations. Flattening them into ordinary triples obscures these distinctions.

Object-Process Methodology (OPM) integrates objects, processes, and states
\cite{iso19450,dori2002opm}. PULSE (Process-aware Unified Language for Semantic
Evolution) adopts that paradigm without claiming an ISO 19450 profile or full
OPM implementation. Executable OPM also predates this work
\cite{yaroker2013opmsim,levisoskin2024maxim}. PULSE instead contributes a
narrower KG-facing role/effect and deterministic spatiotemporal contract, with
OWL/RDF serving as a generated view rather than the authoritative representation.

An OWL--SHACL--workflow composition can reproduce a PULSE trace, as our baseline
demonstrates. We therefore evaluate contract localization: write partitions,
scenario isolation, binding, and temporal order share one checked model/runtime
boundary instead of residing in separately profiled adapters. Observation
acceptance remains at the runner.

The contribution is established by (1) the language and Core calculus with an
explicit write-effect judgment and six safety results; (2) paper proof sketches
and a Lean-checked post-parse subset covering kernel compilation, clocks,
grounded monitors, and selected transition-order fixtures; and (3) standards-composition and Sismic baselines,
mutation analysis, complete modern-era trajectories, and external spatial
engines. Scale and persistence are secondary engineering evidence.

\section{Related Work and Positioning}

OWL 2 provides open-world axioms \cite{motik2012owl}, SHACL validates RDF
graphs \cite{knublauch2017shacl}, GeoSPARQL defines spatial RDF and query
functions \cite{car2024geosparql}, and SOSA/SSN and OWL-Time represent
observations and temporal entities \cite{haller2017ssn,cox2022time}. PULSE
reuses these standards at projection boundaries; none prescribes its four modes
and process runtime.

Reactive semantic data also predates PULSE: RDFTL executes
event-condition-action rules over RDF, C-SPARQL handles continuous queries,
ontology-driven processes infer task executability, and ExeKG translates
KG-described pipelines to scripts
\cite{papamarkos2004rdftl,barbieri2009csparql,rietzke2021execution,zheng2022exekg}.
The PULSE claim is therefore limited to a fixed operational role/effect
partition and clock contract.

PROV-O contextualizes provenance \cite{lebo2013prov}, SHACL-AF adds SPARQL and
triple rules \cite{knublauch2017shaclaf}, and statecharts and Sismic provide mature
event execution \cite{harel1987statecharts,decan2020sismic}. PULSE neither
subsumes their reasoning nor introduces modal logic; it fixes a small write
discipline and sample-and-hold clock across their usual artifact boundaries.

OPM and its RDF exports cover object-process-state modeling
\cite{dori2002opm,iso19450,jacobs2014exporting}; its simulation and MAXIM's
OPCloud extension also make OPM models executable
\cite{yaroker2013opmsim,levisoskin2024maxim}. PULSE instead omits most OPM links,
hierarchy, graphical-textual equivalence, and general computation, selecting
four KG roles, typed write effects, and a fixed spatiotemporal clock.

The Parallel World Framework isolates what-if KG versions through named graphs
and copy-on-write \cite{eibeck2020parallel}; PULSE makes the same
non-interference goal a typed effect rather than an agent/persistence protocol.
LinkML supplies schema-first generators \cite{moxon2026linkml}, while SysML v2
has broader systems scope \cite{omg2026sysml2}; neither prescribes this
role/effect-and-clock contract.

The resulting artifact is a research kernel with a handwritten parser; IDE
support and authoring-cost evaluation remain future work.

\begin{table}[t]
\caption{Where the selected contract resides; ``external'' denotes a supported
composition.}
\label{tab:positioning}
\centering
\scriptsize
\begin{tabularx}{\textwidth}{@{}>{\raggedright\arraybackslash}p{0.19\textwidth}@{\hspace{0.7em}}>{\raggedright\arraybackslash}p{0.25\textwidth}@{\hspace{0.7em}}Y@{}}
\toprule
Approach & Authoritative artifacts & Location of binding, effects, and order \\
\midrule
OWL + SHACL + workflow & Ontology, shapes, engine, adapters & Chosen rule/workflow profile and adapter contract \\
RDF + profiled Sismic & RDF/SHACL, statechart, adapters & Binding load, state invariant, adapter order, clone policy \\
LinkML-assisted stack & Schema and generated artifacts & Generated schema; execution profile remains external \\
OPM/MAXIM & Object-process-state model and OPCloud & Executable model; PULSE omits full OPM and adds KG role/view boundaries \\
SysML v2 & General systems model and tools & Behavior/tool profile; no PULSE-specific role contract \\
PULSE & One typed model and runtime & Compile/runtime checks; evidence acceptance remains external \\
\bottomrule
\end{tabularx}
\end{table}

We operationalize \emph{localization} only as contract placement: its unit is
the declaration or adapter that must change when an obligation changes. The
tested PULSE path edits one typed model consumed by its compiler/runtime,
whereas the profiled baseline places equivalent obligations across RDF/SHACL,
a statechart, and binding, ordering, and clone adapters. Tables \ref{tab:positioning} and
\ref{tab:statechart-faults} record those sites. This structural measure records
placement; authoring time, maintenance cost, and defect rate require separate
study.

\section{Language and Semantics}

\subsection{Modes and Typed Representation}

PULSE co-locates content without assigning one truth status. In this paper,
\emph{mode} denotes an operational role with a distinct write set, and
\emph{normative} denotes guarded constraint validation; neither term asserts a
modal or deontic calculus. Assertions form accepted state, observations are
source-qualified evidence, constraints are read-only obligations, and scenarios
are counterfactual branches. This requirements-derived partition is neither
universal nor exhaustive: merging observations into assertions removes the
authoritative-state non-overwrite boundary, merging constraints into assertions
confuses violations with facts, and merging scenarios into assertions breaks
isolation. Listing \ref{lst:language}
exercises the selected roles with space and time.

Recording evidence leaves transition choice to the runner API\@. PULSE guarantees
authoritative-state non-overwrite before acceptance, while broader evidence
governance remains external. Scenario semantics require source isolation rather
than one copying algorithm: the prototype copies
mutable state, while copy-on-write overlays are equivalent if observably isolated.

\begin{lstlisting}[caption={A compact PULSE spatiotemporal model.},label={lst:language},float=t]
model ColdChainST version "0.1"
crs C = "http://www.opengis.net/def/crs/OGC/1.3/CRS84"
region Z crs C = polygon [[121.49,31.19], [121.51,31.19],
  [121.51,31.21], [121.49,31.21], [121.49,31.19]]
entity Shipment {
  property position: Point crs C
  state condition oneof [Safe, AtRisk]
}
instance batch: Shipment {
  position = point(121.50,31.20) condition = Safe }
observe batch.position = point(121.512,31.201) {
  at "2026-07-19T08:20:00Z" source gps confidence 0.98
  accuracy 5 m }
constraint Containment {
  must coveredBy(batch.position,Z)
  while Shipment.condition == Safe }
process SustainedDeparture(s: Shipment) {
  when leaves(s.position,Z) for 10 min
  changes s.condition: Safe -> AtRisk }
scenario Reroute {
  assume batch.position == point(121.515,31.205)
  run 20 min ask coveredBy(batch.position,Z)
  ask batch.condition }
\end{lstlisting}

The artifact contains this listing as an executable regression case. With no
explicit scenario start, $t_0$ is the latest observation timestamp (or the Unix
epoch if none exists); assumptions execute at $t_0$ in declaration order and
\texttt{run 20 min} advances the isolated branch to $t_0+20$ minutes before the
questions are answered. The example emits the sampled departure, emits its
qualified event at ten minutes, answers \texttt{false} and \texttt{AtRisk}, and
leaves the source model unchanged.

The immutable document is
$M=\langle n,v,R,E,I,O,C,P,S\rangle$. Table \ref{tab:ir} records the principal
implemented fields; optional values are explicit. Parsing is followed by name,
type, state-domain, geometry, CRS, and reference validation before a runtime
world can be created.

\begin{table}[t]
\caption{Principal records in the implemented typed representation.}
\label{tab:ir}
\centering
\scriptsize
\begin{tabularx}{\textwidth}{@{}>{\raggedright\arraybackslash}p{0.15\textwidth}@{\hspace{0.6em}}Y@{\hspace{0.6em}}>{\raggedright\arraybackslash}p{0.23\textwidth}@{}}
\toprule
Record & Fields & Invariant \\
\midrule
Model & \texttt{name, version, crs*, regions*, entities*, instances*, observations*, constraints*, processes*, scenarios*} & Declaration names resolve uniquely \\
Geometry & \texttt{kind, coordinates, crs} & Finite coordinates; explicit CRS; simple closed polygon shell \\
Observation & \texttt{subject, feature, value, at, source, confidence?, accuracy?} & Time has offset; quality values bounded \\
Constraint & \texttt{name, predicate, subject, region, while?} & Types and CRS agree \\
Process & \texttt{name, parameter, event, subject, region, duration?, transition} & State source/target declared \\
Scenario & \texttt{name, assumptions*, runFor?, questions*} & Assumptions type-check; source world immutable \\
\bottomrule
\end{tabularx}
\end{table}

\subsection{Spatial and Temporal Execution}

Execution uses configurations
$X=\langle A,O,q,Q,t\rangle$: asserted values $A$, append-only evidence $O$,
finite object states $q$, a finite partial timer map $Q$, and event time $t$.
The compiler grounds a parameterized process once per compatible instance, in
instance-declaration order, assigning identifiers such as
\texttt{SustainedDeparture@batch}. $Q$ is keyed by that ground identifier, so
different subjects may hold concurrent monitors for one source process. The
declaration \texttt{observe} performs only
$\mathsf{record}(o):\langle A,O,q,Q,t\rangle\mapsto
\langle A,O\mathbin{+}o,q,Q,t\rangle$. Authoritative change is instead the
runner action $\mathsf{move\_at}(i,p,t')$; it is an input command, not a model
declaration. The application owns the acceptance decision that may cause this
command; the language guarantees only that recording evidence alone never
triggers a process.

Let $\mathsf{in}_c(g,r)$ be boundary-inclusive Point/Polygon membership under
CRS $c$. Before a move, $t'\geq t$, references, and CRS equality are checked;
failure has no effect. For every accepted position after the first, including
$t_{k-1}=t_k$, let $g_{k-1}$ be the immediately preceding accepted sample and
$g_k$ the new one. Sampled entry/exit is derived by
\[
\begin{split}
\mathsf{enters}(i,r,t_k)&=\neg\mathsf{in}(g_{k-1},r)\land
                              \mathsf{in}(g_k,r),\\
\mathsf{leaves}(i,r,t_k)&=\mathsf{in}(g_{k-1},r)\land
                              \neg\mathsf{in}(g_k,r).
\end{split}
\]
The runner first emits timers due by $t'$, ordered by
$(\mathit{deadline},\mathit{groundRank})$; it
then compares old/new membership, updates $A$, cancels inverse monitors, starts
new monitors, and applies immediate rules in declaration order. Ground rank is
assigned by process then instance declaration, so alpha-renaming is inert. Later rules see
earlier state mutations; a source-state mismatch skips a rule. Same-time input
moves follow input order and may therefore create crossings and cancel/start
monitors, but no time elapses between them. Positions are sample-and-hold: an outside-to-outside
segment that intersects a polygon yields no sampled event. A monitor
$\langle e,t_0,\delta\rangle$ cancelled before $t_0+\delta$ emits nothing;
otherwise its effective time is $t_0+\delta$ and its emission time is the first
clock advance exposing the deadline. Thus a due timer precedes a same-time
move. A scenario may clone an active $X$, including its clock and pending
monitors; the default CLI branch instead starts from the compiled base world
with $Q=\emptyset$. Both execute assumptions as moves only on the clone.

These write sets induce four implementation invariants. \emph{P1}
$\mathsf{record}$ cannot overwrite $A$. \emph{P2} scenario effects cannot
reach the source configuration. \emph{P3} validation plus guarded rules keeps
$q$ within declared state domains. \emph{P4} fixed validated input, timer
ordering, input order, and rule order yield one trace or an atomic error. The
next section states these properties over a core calculus.

\section{Core Calculus and Properties}

The surface language is resolved and desugared to a finite kernel. Let $I$ and
$S_i$ be finite sets of instances and states, and let $R$ and $K$ be
declaration-ordered sequences of regions and ground specification identifiers.
Each parameterized process is expanded over compatible instances to a unique
$k\in K$ with rank $\rho(k)$. For each CRS $c$, $G_c$ is the domain of finite valid
geometries in $c$. The implemented fragment restricts authoritative moving
values to points and regions to simple polygons. The immutable environment is
\[
 \Gamma=\langle\mathit{crs},\mathit{region},\mathit{stateDomain},
 \mathit{rules},\mathit{specs},\mathit{constraints}\rangle .
\]
Rules and sustained-event specifications are declaration-ordered; ground
identifiers are unique and durations are positive.
Following standard type-safety structure, the relevant
judgments are $\Gamma\vdash g:\mathsf{Geometry}[c]$,
$\Gamma\vdash o:\mathsf{Observation}[i]$, $\Gamma\vdash a:\mathsf{Action}$,
and $\Gamma\vdash X\;\mathsf{ok}$. A spatial predicate is well typed only when
its point and polygon share one CRS.

Core actions and outcomes are
\[
\begin{split}
a &::= \mathsf{record}(o)\mid \mathsf{move}(i,g,t')
       \mid \mathsf{advance}(t'),\\
u &::= \mathsf{ok}(X',T)\mid \mathsf{error}(e,X),
\end{split}
\]
where $T$ is a finite ordered trace. Well-formedness requires resolved
identifiers, declared CRSs, $q(i)\in S_i$, typed offset-aware observations,
unique pending monitors, and offset-aware time. A pending monitor contains a
ground identifier, event, subject, region, start, positive duration, and an
optional guarded rule.

The write-effect judgment $\Gamma\vdash a\triangleright W$ assigns
$\mathsf{record}\mapsto\{O\}$,
$\mathsf{move}\mapsto\{A,q,Q,t\}$, and
$\mathsf{advance}\mapsto\{q,Q,t\}$; validation and questions have
$W=\emptyset$. A successful action leaves every configuration component
outside $W$ unchanged. This is the limited sense in which PULSE is an effect
discipline; the judgment is paper-level bookkeeping, not a separately
mechanized general effect system.

Evaluation is a deterministic big-step relation
$\Gamma\vdash\langle X,a\rangle\Downarrow u$. Observation recording has the
single rule
\[
\frac{\Gamma\vdash o:\mathsf{Observation}}
 {\Gamma\vdash\langle\langle A,O,q,Q,t\rangle,\mathsf{record}(o)\rangle
  \Downarrow\mathsf{ok}(\langle A,O{\cdot}o,q,Q,t\rangle,\epsilon)}.
\]
For time advance, $\mathsf{due}(Q,t')$ selects deadlines no later than $t'$ and
sorts them by $(\mathit{deadline},\rho(k))$; unique ground keys make equal
deadlines unambiguous without depending on spelling. $\mathsf{emit}$ removes those monitors,
emits their semantic deadline and current emission time, and applies an
attached rule only when its source state still matches:
\[
\frac{t'\geq t\quad D=\mathsf{due}(Q,t')\quad
      \mathsf{emit}(X,D,t')=\langle A,O,q',Q',T\rangle}
 {\Gamma\vdash\langle X,\mathsf{advance}(t')\rangle
  \Downarrow\mathsf{ok}(\langle A,O,q',Q',t'\rangle,T)}.
\]

A move validates all premises before mutation, advances time, derives sampled
crossings, commits the position, updates monitors, and applies immediate rules:
\[
\frac{\begin{gathered}
 t'\geq t\quad\Gamma\vdash g:\mathsf{Point}[\mathit{crs}(i)]\\
\Gamma\vdash\langle X,\mathsf{advance}(t')\rangle
 \Downarrow\mathsf{ok}(X_0,T_d)\\
E=\mathsf{cross}(A_0,i,g)\quad A_1=A_0[i\mapsto g]\quad
Q_1=\mathsf{cancelStart}(q_0,Q_0,E,t')\\
q_1=\mathsf{applyRules}(q_0,E)\quad
X_1=\langle A_1,O_0,q_1,Q_1,t'\rangle
\end{gathered}}
{\begin{gathered}
\Gamma\vdash\langle X,\mathsf{move}(i,g,t')\rangle\\[-1mm]
\Downarrow\mathsf{ok}(X_1,T_d{\cdot}E)
\end{gathered}}.
\]
The functional validator $\mathsf{validate}(\Gamma,X,a)$ checks the timestamp
form, resolved identifier, time monotonicity, then type/CRS compatibility and returns the
first error or $\mathsf{ok}$; the single error rule returns
$\mathsf{error}(e,X)$. Monitor creation
requires the attached rule's source-state guard
to hold in $q_0$; emission rechecks the guard before applying its transition.
Scenarios evaluate a finite desugared action sequence over a fresh branch value
$X_s=\mathsf{clone}(X)$. The start is
$t_0=\max(X.t,\mathsf{latest}(O),t_{\mathrm{explicit}})$ when an explicit
start exists, omitting absent terms. The clone first advances to $t_0$, then
applies assumptions there; a declared horizon $d$
desugars to $\mathsf{advance}(t_0+d)$. The calculus treats $X_s$ as a fresh
value; the Python implementation's no-alias copy is checked separately.
Normative validation and scenario questions are read-only judgments
$\Gamma\vdash X\Downarrow_N V$ and
$\Gamma\vdash\langle X_s,\overline{q}\rangle\Downarrow_Q\overline{b}$.
$V$ is the declaration-ordered sequence of active constraints whose predicate
is false (a false optional state guard makes a constraint inactive), while
$\overline{b}$ answers questions in declaration order after assumptions and
the horizon advance. Both judgments leave their input configuration unchanged;
the Python implementation executes them, but they are not core mutation actions
or part of the current Lean-checked subset.

\subsection{Core Properties}

\textbf{Lemma 1 (helper and effect confinement).} On well-formed input,
$\mathsf{cross}$ traverses $R$ in declaration order; $\mathsf{cancelStart}$
traverses that event sequence and $K$ in rank order, preserving ground-key
uniqueness and future deadlines; $\mathsf{due}/\mathsf{emit}$ use the total key
$(\mathit{deadline},\rho(k))$; and $\mathsf{applyRules}$ preserves state domains.
Each helper has one finite result, and successful actions mutate only the
components in their declared $W$.

\textbf{Theorem 1 (preservation).} If $\Gamma\vdash X\;\mathsf{ok}$,
$\Gamma\vdash a:\mathsf{Action}$, and
$\Gamma\vdash\langle X,a\rangle\Downarrow\mathsf{ok}(X',T)$, then
$\Gamma\vdash X'\;\mathsf{ok}$.
\emph{Proof.} By cases on $a$. Record appends a typed observation. Advance
only removes well-formed monitors and assigns declared rule targets. Move
validates identifier, time, and CRS before assigning a typed point; new
monitors originate in $\Gamma$, cancellation only removes entries, and rules
assign declared targets. Unchanged components inherit their invariant.

\textbf{Theorem 2 (determinism).} If
$\Gamma\vdash\langle X,a\rangle\Downarrow u_1$ and
$\Gamma\vdash\langle X,a\rangle\Downarrow u_2$, then $u_1=u_2$.
\emph{Proof.} Validation is functional. In successful cases Lemma 1 fixes
each intermediate result, and every order-sensitive operation has an explicit
order. The error rule retains $X$ and returns the unique validation error.

\textbf{Theorem 3 (record confinement).} If recording $o$ maps $X$
to $X'$, then $X'.A=X.A$, $X'.q=X.q$, $X'.Q=X.Q$, and $X'.t=X.t$.
This is immediate from the recording rule. Appended evidence can still affect
later operations: for example,
$\mathsf{latest}(O)$ selects a default scenario start.

\textbf{Theorem 4 (extensional scenario isolation).} Scenario evaluation leaves
every component of source $X$ unchanged because core steps consume and return
only the fresh branch value. Python heap-alias absence is tested separately at
the implementation level.

\textbf{Theorem 5 (finite advance).} For finite $Q$ and $t'\geq t$,
$\mathsf{advance}(t')$ terminates and emits at most $|Q|$ sustained events.
The due set is finite; emission removes one distinct keyed monitor per
iteration and starts none.

\textbf{Theorem 6 (atomic failure).} If evaluation returns
$\mathsf{error}(e,X_e)$, then $X_e=X$. This follows directly from validation
before mutation and the error rule.

\textbf{Conditional projection adequacy.} Assume a GeoSPARQL processor uses
the standard interpretation of CRS84 WKT Point/Polygon literals. For the
supported fragment, strict \texttt{inside} agrees with \texttt{sfWithin}, and
boundary-inclusive \texttt{coveredBy} agrees with \texttt{sfIntersects}.
The latter mapping is fragment-specific: it relies on a Point subject and
Polygon region. Section~\ref{sec:evaluation} tests the condition with an
unmodified external engine. Agreement checks this fragment; formal conformance
requires the official test suite.

\section{Semantic Views and Prototype}

The Python prototype contains an EBNF parser, immutable typed records, semantic
compiler, dependency-free Point/Polygon kernel, role-aware runtime, temporal event
queue, projection layer, and command-line tools. Table \ref{tab:views} makes the
responsibility split explicit. Execution stays in the typed representation;
the views are purpose-specific and provide partial, role-dependent fidelity.
Table \ref{tab:views} is also a fidelity contract: asserted geometry and
observations use selected vocabularies; processes and scenarios have JSON-only
outputs; runtime ordering, queues, and branch identity are not reconstructible.
The implementation supports forward generation only.

\begin{table}[t]
\caption{Construct coverage, projection fidelity, and responsibility boundary.}
\label{tab:views}
\centering
\scriptsize
\begin{tabularx}{\textwidth}{@{}>{\raggedright\arraybackslash}p{0.14\textwidth}@{\hspace{0.5em}}>{\raggedright\arraybackslash}p{0.23\textwidth}@{\hspace{0.5em}}>{\raggedright\arraybackslash}p{0.22\textwidth}@{\hspace{0.5em}}Y@{}}
\toprule
Surface role & Runtime/Core & KG view & Checked or external boundary \\
\midrule
Assertion, move & $A$; writes $A,q,Q,t$ & RDF/GeoSPARQL & Core/Integrated Lean; evidence acceptance external \\
Observation & $\mathsf{record}$ writes only $O$ & SOSA plus source/quality & Core Lean record confinement; promotion external \\
Constraint & Read-only $\Downarrow_N$ & SHACL-SPARQL & Python/pySHACL; not Lean or a transition \\
Duration, process & Grounded $q,Q,t$ rules & JSON event; no RDF view & Integrated Lean; geometry abstracted \\
Scenario, query & Branch actions; read-only $\Downarrow_Q$ & JSON answer; no RDF view & Compiler horizon and extensional isolation; questions not Lean \\
Compiler, view & Resolve, type-check, generate & Standards views & Regression checks; no surface soundness or general refinement theorem \\
\bottomrule
\end{tabularx}
\end{table}

The optional projection validator executes generated SHACL with pySHACL and a
GEOS adapter for the referenced GeoSPARQL functions. It compares conformance and
violations with internal validation. This detects cross-view drift; GeoSPARQL
server behavior and conformance remain external. A separate container pins
Apache Jena GeoSPARQL 6.1.0 \cite{osman2022geosparqljena}. Its Java harness
imports no PULSE geometry code, loads a projected Turtle graph, evaluates
standard \texttt{sfWithin}, \texttt{sfIntersects}, \texttt{sfDisjoint}, and
\texttt{sfTouches} functions, and emits SPARQL Results JSON plus separately
measured initialization, RDF-load, and query-materialization times. Two
separately enabled profile modules fill the GeoSPARQL 1.1 Geometry surface and
an H3 4.4.0 DGGS surface; they are measured both apart from and
in combination with the unmodified Jena baseline.

\section{Evaluation}
\label{sec:evaluation}

The evaluation follows the contribution hierarchy. Its primary questions are
whether the contract survives executable and reduced mechanized checks (RQ1), whether
its spatial projections agree with external engines (RQ2), whether traces remain
stable on complete modern-era data
(RQ3), and whether finite exhaustive temporal traces and single-change mutants make the
operational commitments observable in PULSE and a reference workflow (RQ4).
Scale and persistent-index checks assess
component behavior (RQ5--RQ6). A durable PostGIS workload is reported only as
a secondary feasibility stress test (RQ7); modeling ease and end-to-end database
behavior lie outside its measurements. Table
\ref{tab:evidence} summarizes these deliberately bounded claims.

\begin{table}[H]
\caption{Executed evidence and deliberately bounded claims.}
\label{tab:evidence}
\centering
\scriptsize
\begin{tabularx}{\textwidth}{@{}>{\raggedright\arraybackslash}p{0.16\textwidth}@{\hspace{0.8em}}>{\raggedright\arraybackslash}p{0.37\textwidth}@{\hspace{0.8em}}Y@{}}
\toprule
Evidence & Workload and reference & Result and boundary \\
\midrule
Core properties & 88 tests; Lean 4; 340 bounded move traces; integrated compiler/transition/monitor model & Listing, live-branch cloning, identifier rejection, and two-subject grounding execute; IR plus 32 runtime-kernel projections agree; 3,534 checks; no failures \\
Temporal sensitivity & 37,440 generated traces; ten single-field reference-workflow mutants & PULSE/reference 37,440/37,440; mutants killed 10/10 (1,680--11,024 distinguishing traces) \\
Spatial agreement & 89-case GEOS corpus; shared $86\times86$ Jena/PostGIS graph & 0 GEOS differences; both external engines return 7,396 rows without differences \\
GeoSPARQL inventory & 55 Annex identifiers; 185 custom probes; official RDF source audit & Native Jena 112/185; PULSE profiles 185/185; custom coverage, not conformance \\
IBTrACS scale & 4,775 tracks; 1,476,290 transition-zone pairs & 4,800 sampled; 12,831 sustained; 0 differences; no latitude-band seam artifact \\
Contract-location composition & 91 samples; three cold-chain encodings; Sismic fault-location study & Exact traces agree; six contract-site faults located across three configurations \\
Secondary components & $10^5$ scale ladder; PostGIS GiST, restart, and load probes & Counts preserved; persistence feasible; load results are not PULSE throughput \\
\bottomrule
\end{tabularx}
\end{table}

\subsection{Core Contract and Policy Checks (RQ1)}

The 88 tests exercise role, CRS, state, parser, projection, SHACL, timer,
compiler, and multi-subject contracts, including Listing \ref{lst:language}.
A four-position finite abstraction enumerates 340 move traces through depth
four and exposes pending deadlines; all 3,534 determinism, preservation,
finite-advance, atomicity, and isolation checks pass within this bounded
abstraction.

A Lean 4.30.0 development \cite{demoura2021lean4}, with no \texttt{sorry},
checks functional kernels for positions, evidence, clocks, monitors, atomicity,
and scenarios. Conditional compiler lemmas preserve guards, triggers,
durations, deadlines, horizons, and actions; Lean and Python emit byte-identical
IR for Listing \ref{lst:language}.

Six named \path{Core} declarations cover analogues of Theorems 1--6. The
integrated model checks both rule classes, due-before-crossing order, guarded
updates, and atomic time/CRS failure, but neither object-state preservation nor
general surface compilation is a Lean theorem. A 32-case bridge agrees on final
state, pending count, and ordered event fields; declaration order is checked by
an alpha-renaming regression and a two-symbol Lean prefix. Geometry, parsing,
questions, and general Python refinement remain unmechanized. Python tests also
reject undeclared subjects atomically and clone live scenario state without aliases.

Six policy-sensitivity cases each replace one declared contract; every
alternative changes state or event trace (6/6). RQ4 instead mutates one workflow
switch at a time. These results make the commitments observable; policy
preference remains unevaluated.

\subsection{External Spatial and Projection Agreement (RQ2)}

A differential corpus compares \texttt{within}, \texttt{onBoundary}, and
\texttt{coveredBy} with GEOS across edges, vertices, concavity, ring
orientation, near-boundary points, thin polygons, large offsets, and four CRS84
translations/scales. All 89 valid GEOS cases matched; 86 CRS84 cases feed the
external-engine comparison, and nine malformed geometry/CRS cases were rejected.
The adaptive exact-orientation fallback avoids a fixed
coordinate-unit epsilon. These are project-level differential tests, distinct
from the official GeoSPARQL suite.

Generated RDF and SPARQL parse independently, and pySHACL and GEOS agree with
internal constraints on four boundary and guard conditions. Apache Jena then
evaluated all 7,396 pairs of the 86 projected points and regions using four
Simple Features predicates; every row matched PULSE, including 38 actual
point-on-shell self-pairs. PostGIS returned the same rows without differences. This triangulation
also corrected the projection: \texttt{ehCoveredBy} does not capture PULSE's
boundary-inclusive Point/Polygon membership, whereas \texttt{sfIntersects}
does for this fragment.

\subsection{Secondary GeoSPARQL Profile Probes (RQ2)}

As secondary interface coverage, a project manifest maps 55
GeoSPARQL 1.1 Annex A identifiers \cite{car2024geosparql} to 185 custom probes.
Native Jena passes 112/185; the isolated Geometry and H3 profiles pass 185/185.
A pinned source audit and per-group purposes are preserved in the artifact.
These custom probes complement rather than execute the OGC abstract tests;
conformance remains outside their scope.

\subsection{Real Trajectories and Duration Semantics (RQ3)}

The frozen NOAA IBTrACS v04r01 \texttt{since1980} CSV
\cite{knapp2010ibtracs} contains 307,382 rows (143 MB). Valid main-track rows
yield 4,775 tracks, 300,033 points, and 295,258 transitions across seven basins;
4,768 tracks have transitions. A single-zone replay matched GEOS on every
transition, including 571 event-bearing transitions from 501 tracks.

Five-zone replay compared every membership, crossing, and 6/12/24-hour event
with GEOS and an event sweep implemented separately by the first author. All
1,476,290 pairs matched: 4,800 sampled events and 12,831 emissions from 14,400
monitor starts. A dateline audit retained all tracks and found 420 normalized
longitude jumps in 366 tracks across the Eastern, Southern, and Western Pacific.
After global latitude bands were closed at $\pm180^\circ$, none of their
membership changes was seam-only; 108 changes remained at the deliberately
bounded $179.999^\circ$E edge of the Western-Pacific study zone. Reported events
therefore use sample-and-hold semantics; continuous-segment and
antimeridian-crossing polygon interpretations remain outside this experiment.

\subsection{End-to-End Execution and Composition (RQ4)}

A 91-point track exercised all four operational roles: 90 accepted moves produced three
sampled events, one six-hour event, one
\texttt{Safe}$\rightarrow$\texttt{AtRisk} change, and two guarded violations.
Observation recording and a hypothetical move preserved asserted source state.
The 926 data and 6 shape triples parsed, and pySHACL and GEOS matched internal
validation. This single integration case assesses execution coherence;
deployment behavior remains outside its scope.

A cold-chain trace was also implemented in PULSE, a workflow composed from
GeoSPARQL, SOSA, OWL-Time, and SHACL, and an OGC MF-JSON \texttt{Step}
workflow. All reproduced cancellation, a ten-minute departure, and final state
\texttt{AtRisk}. Because the composed paths share reference workflow machinery
and adapters, we exclude source-line and file counts from modeling-effort and
usability claims.

We then retained the RDF/SHACL inputs but replaced the reference workflow with
Sismic 1.6.11 \cite{decan2020sismic}. The statechart, implemented separately by
the first author,
matched PULSE exactly on final state, all three crossings, cancellation, and
the sustained event's start, effective, and emission timestamps. Table
\ref{tab:statechart-faults} injects one fault at a time. The six faults were
selected to exercise distinct contract sites---identifier binding, effect
domain, sample/clock order, scenario isolation, evidence/source role, and
monitor start guard. The unprofiled
composition passes static graph validation and needs a complete outcome/source
oracle; adding an RDF--statechart binding check, a state invariant, and adapter
preconditions moves detection earlier. All statecharts, workflows, and adapters
were implemented by the first author. The second author independently reviewed
and validated every baseline implementation and reported comparison. The review
audits existing implementations; independent reimplementation remains future
work. The case localizes contract sites on one task, leaving authoring cost and
generality open.

\begin{table}[t]
\caption{Detection stage for matched single-site integration faults. ``Oracle''
means comparison with the unchanged complete outcome/source state.}
\label{tab:statechart-faults}
\centering
\scriptsize
\begin{tabularx}{\textwidth}{@{}Y@{\hspace{0.6em}}>{\raggedright\arraybackslash}p{0.16\textwidth}@{\hspace{0.6em}}>{\raggedright\arraybackslash}p{0.16\textwidth}@{\hspace{0.6em}}>{\raggedright\arraybackslash}p{0.22\textwidth}@{}}
\toprule
Injected fault & PULSE & RDF+Sismic & Profiled RDF+Sismic \\
\midrule
Region identifier drift & compile & oracle & binding load \\
Effect outside state domain & compile & oracle & invariant \\
Sample processed before clock & runtime API & oracle & adapter precondition \\
Scenario aliases source state & scenario clone & source oracle & clone adapter \\
Observation overwrites source & record API & source oracle & role adapter \\
Guard-false monitor start & runtime guard & oracle & adapter precondition \\
\bottomrule
\end{tabularx}
\end{table}

To test temporal obligations without assigning workflow-code defects to SHACL,
we generated every trace of lengths two through five over Boolean membership,
increments of 1, 5, 10, or 11 minutes, initial states \texttt{Safe} or
\texttt{Maintenance}, and presence/absence of a same-trigger immediate
transition. This Cartesian product contains
$4\sum_{n=2}^{5}2^n4^{n-1}=37{,}440$ traces. For every trace the PULSE runtime
and a reference workflow implemented in separate code by the first author
agreed exactly on final state, instantaneous events, start, effective, and
emission timestamps for sustained events, and ordering.

The reference workflow exposes a declared nine-switch experimental mutation
model. We changed each switch alone and used two directional substitutions for duration scale, producing ten
operators: inverse cancellation; timer-before-move; start and deadline guards;
deadline equality; shorter and longer duration; emission timestamp;
transition-on-start; and pre- versus post-immediate monitor eligibility. A
runtime assertion rejects any operator that changes more than one field or any
field without an operator. All ten mutants were killed; depending on the
operator, 1,680--11,024 generated traces distinguished it, and the report stores
the first exact witness. Workflow-fault detection uses the outcome oracle and
excludes RDF or SHACL results. Exhaustiveness applies to the declared grid; the
operator schema remains researcher-defined.

The baselines confirm that each composition computes the tested trace. Their
purpose is to identify where the obligations reside: PULSE validates
names, types, CRS, operational roles, and runtime effects inside one boundary; the profiled
statechart composition recovers all six protections only after binding,
state-invariant, adapter-order, clone-isolation, role, and start-guard checks are added beside its
RDF, SHACL, and statechart artifacts. In this task, unprofiled composition finds
6/6 faults only at the complete outcome/source oracle; PULSE rejects two at
compile time and prevents four at its runtime, record, or scenario boundary.
These results characterize contract location and maintenance surface; modeling
effort, usability, and language-level superiority require separate studies.

\subsection{Secondary Engineering Checks (RQ5--RQ7)}

Artifact-only checks preserve counts through $10^5$ moves, retain 300,033
points and GiST indexes across PostgreSQL/PostGIS container replacement, and
reproduce the IBTrACS sweep from stored memberships. A
50,000-object read/update/event workload also survives crash/restart; its
open-loop admission boundary is host-specific. These checks cover component
persistence and host-specific operation while bypassing parts of PULSE.

\section{Discussion and Limitations}

Executable OPM, reactive RDF rules, ontology-driven execution, and isolated KG
scenarios predate PULSE; a standards/workflow composition reproduces the tested
trace. PULSE contributes placement, not new computational power: one typed
runtime localizes evidence, obligations, hypothetical effects, and timer/state
ordering, while acceptance remains external.

The mutation and statechart executions use the same complete-loop outcome as
PULSE, with workflow faults assessed by the outcome oracle rather than SHACL\@.
Finite enumeration avoids trace cherry-picking, while the nine switches and six
contract-site faults remain a researcher-defined sample. The first author
implemented every executable path; the second author independently reviewed and
validated the baseline implementations and reported comparisons. Agreement plus
this audit supports cross-implementation consistency. Independent
reimplementation of the intended semantics remains future work.

The four roles form an intentionally non-exhaustive, requirements-derived
partition. PULSE covers pre-acceptance authoritative-state non-overwrite and
branch isolation; acceptance policy and copy/overlay storage remain external.

The result boundary is correspondingly narrow. Geometry covers points and
simple polygons under planar predicates and explicit CRS assumptions; holes,
multipolygons, transformations, geodesics, uncertainty, continuous crossings,
and antimeridian-crossing polygons remain unsupported. The custom GeoSPARQL
probes measure interface coverage, while official OGC conformance remains
untested. H3 is approximate. Single-node PostGIS measurements cover component
persistence and load, leaving end-to-end throughput, failover, and a portable
SLA open. IBTrACS is retrospective, the contract-location case uses one track,
and authoring-cost claims await a user study.
Lean starts after parsing and abstracts geometry, floating-point predicates,
constraint validation, and scenario questions. Its 32-case bridge also omits
specification name, start, and duration, providing case correspondence rather
than general Python refinement. Exhaustive mutation is bounded by its declared
grid.

Future work should independently reimplement the profiles, add interpolation,
broaden Python--Lean refinement, test replicated deployments, and preregister a
modeling study against composed standards and schema-first baselines.

\section{Conclusion}

PULSE contributes a typed executable contract for asserted state, evidence,
obligations, counterfactuals, and clocked spatial processes. Standards/Sismic
baselines reproduce the tested protections through explicit cross-artifact
contracts. On a finite grid, 37,440 exact matches and ten killed mutants expose
temporal commitments. The calculus, Lean subset, IBTrACS replay, and external
spatial checks support safety and trace parity for the implemented fragment;
persistence and load tests remain secondary component results.

\paragraph{Artifact Availability.}
Grammar, source, tests, data, protocols, Lean files, and reports are in the
\href{https://github.com/deeplethe/pulse-spatial}{PULSE spatial artifact}, commit
\texttt{798fb7e4e4ef4c04318f0790bae99bc191802e81}.

\paragraph{Author Contributions.}
Dongxu Yang conceived PULSE, designed the language and formal semantics,
implemented the compiler and runtime, conducted the experiments, and drafted
the manuscript. Ziyi Liang independently reviewed and validated all baseline
implementations and reported comparisons. Both authors accept responsibility
for the manuscript.

\paragraph{Declaration of Generative AI Use.}
Generative AI tools were used for language editing, citation and formatting
assistance, reviewer-style critique, and test scaffolding. Dongxu Yang
independently originated the research problem and PULSE's novel contribution;
drafted the manuscript; designed the language, formal semantics, theorem
statements, proofs, experimental protocol, and evaluation claims; designed and
implemented the prototype, baselines, and experiments; interpreted the results;
and verified the assisted material, citations, code, proofs, data, and reported
results. Both authors accept full responsibility for the content.

\bibliographystyle{abbrv}
\bibliography{references}

\end{document}